\def\year{2022}\relax
\documentclass[letterpaper]{article} 
\usepackage{aaai22}  
\usepackage{times}  
\usepackage{helvet}  
\usepackage{courier}  
\usepackage[hyphens]{url}  
\usepackage{graphicx} 
\usepackage{natbib}  
\usepackage{caption} 
\DeclareCaptionStyle{ruled}{labelfont=normalfont,labelsep=colon,strut=off} 
\usepackage{algorithm}
\usepackage{algorithmic}

\usepackage[utf8]{inputenc} 
\usepackage{url}            
\usepackage{booktabs}       
\usepackage{amsfonts}       
\usepackage{nicefrac}       
\usepackage{microtype}      
\usepackage[dvipsnames]{xcolor}

\title{Analytic Deterministic Policy Gradient with Application to Datacenter Congestion Control}

\usepackage{graphicx}
\usepackage{array}
\usepackage{amsthm}
\usepackage{amsmath}
\usepackage{multirow}
\usepackage{diagbox}

\usepackage{verbatim}
\usepackage[colorinlistoftodos]{todonotes}

\DeclareMathOperator*{\argmax}{arg\,max}

\DeclareMathOperator{\bigO}{\mathcal{O}}

\DeclareMathOperator{\action}{\mathbf{a}}
\DeclareMathOperator{\state}{\mathbf{s}}

\DeclareMathOperator{\E}{\mathbb{E}} %

\usepackage{caption}
\usepackage{subcaption}
\usepackage{sidecap}

\usepackage{thm-restate}
\usepackage{bibunits}
\usepackage[capitalise]{cleveref}
\usepackage{colortbl}
\usepackage{float}

\newtheorem{proposition}{Proposition}

\newtheorem{claim}[proposition]{Claim}
\newtheorem{definition}{Definition}

\newcommand{\failed}{{packet loss}}

\def\E{{\mathbb{E}}}

\def\alg{{ADPG}}
\def\algfull{{Analytic Deterministic Policy Gradient}}

\newcommand\numberthis{\addtocounter{equation}{1}\tag{\theequation}}

\newcommand{\ignore}[1]{}

\usepackage{pifont} 
\newcommand{\cmark}{\ding{51}}%
\newcommand{\xmark}{\ding{55}}%

\definecolor{mygreen}{RGB}{62, 150, 81}

\usepackage{newfloat}
\usepackage{listings}
\floatstyle{ruled}
\newfloat{listing}{tb}{lst}{}
\floatname{listing}{Listing}
\title{Reinforcement Learning for Datacenter Congestion Control}

\author {
    Chen Tessler\textsuperscript{\rm 1 \rm 2}\thanks{Work done during an internship at Nvidia Research.}
    Yuval Shpigelman\textsuperscript{\rm 3}
    Gal Dalal\textsuperscript{\rm 2}
    Amit Mandelbaum\textsuperscript{\rm 3}
    Doron Haritan Kazakov\textsuperscript{\rm 3}
    Benjamin Fuhrer\textsuperscript{\rm 3}
    Gal Chechik\textsuperscript{\rm 2 \rm 4}
    Shie Mannor\textsuperscript{\rm 1 \rm 2}

}
\affiliations {
    \textsuperscript{\rm 1} Technion Institute of Technology,
    \textsuperscript{\rm 2} Nvidia Research,
    \textsuperscript{\rm 3} Nvidia Networking,
    \textsuperscript{\rm 4} Bar-Ilan University\\
    chen.tessler@gmail.com, \{yuvals,gdalal,amitma,doronh,bfuhrer,gchechik,smannor\}@nvidia.com
}

\begin{document}

\maketitle

\begin{abstract}
    We approach the task of network congestion control in datacenters using Reinforcement Learning (RL). Successful congestion control algorithms can dramatically improve latency and overall network throughput. Until today, no such learning-based algorithms have shown practical potential in this domain. Evidently, the most popular recent deployments rely on rule-based heuristics that are tested on a predetermined set of benchmarks. Consequently, these heuristics do not generalize well to newly-seen scenarios. Contrarily, we devise an RL-based algorithm with the aim of generalizing to different configurations of real-world datacenter networks. We overcome challenges such as partial-observability, non-stationarity, and multi-objectiveness. We further propose a policy gradient algorithm that leverages the analytical structure of the reward function to approximate its derivative and improve stability. We show that these challenges prevent standard RL algorithms from operating within this domain. Our experiments, conducted on a realistic simulator that emulates communication networks' behavior, show that our method exhibits improved performance concurrently on the multiple considered metrics compared to the popular algorithms deployed today in real datacenters. Our algorithm is being productized to replace heuristics in some of the largest datacenters in the world.
\end{abstract}

\section{Introduction}

Modern datacenters consist of multiple servers jointly communicating at ultra high speeds. As the joint transmission rate of the servers surpasses the internal connection limitations (e.g., router processing rate) the communication may become congested. Congested networks suffer from reduced bandwidth utilization, the appearance of packet loss and increased application latency. Hence, avoiding and preventing congestion is an important task called congestion control.

Previous work on congestion control has mainly focused on rule-based methods. As these schemes are rule based, they are usually optimized for a single set of tasks (as we show in \cref{table: comparison}). Machine learning methods, as opposed to rule-based ones, are capable of learning and generalizing based on data and experience. Specifically, reinforcement learning (RL) automatically learns a control policy (transmission rate control in the context of congestion prevention) given an environment to interact with and a reward signal. Thus, an RL algorithm is (1) capable of solving the task without tedious manual tuning of control parameters and (2) successfully operate in a vast set of tasks, e.g., generalize, when provided with diverse training environments.

Most RL algorithms were designed under the assumption that the world can be adequately modeled as a Markov Decision Process (MDP). Unfortunately, this is seldom the case in realistic applications and specifically datacenter congestion control. As we show below, as opposed to the standard assumptions, this problem is partially-observable and consists of multiple-agents. These challenges prevent the standard RL methods from coping within such a complex environment, and we believe are one of the major reasons such methods are yet to be deployed in real world datacenters.

From an RL point of view, each agent controls the transmission of a single application. As there are multiple applications across multiple servers, this means multiple agents that, due to security reasons, are unaware of one another and unable to communicate -- hence multi-agent and partially observable. We observe that popular RL algorithms such as DQN \citep{mnih2015human}, REINFORCE \citep{williams1992simple} and PPO \citep{schulman2017proximal} fail on such tasks (\cref{tab: many-to-one comparison}).

To overcome these challenges, we present the \algfull{} (\alg{}), a scheme that makes use of domain knowledge to estimate the gradient for a deterministic policy update. As this task lacks a ground truth reward function, we present a fitting reward function and show that this reward function, in a multi-agent partially-observable setting, leads to convergence to a global optimum.

\begin{table*}
\begin{center}
\begin{tabular}{c|cccc}
     & \textbf{Many to One} & \textbf{All to All} & \textbf{Long Short} & \textbf{Real World} \\
    \midrule
    Aurora / PPO / REINFORCE & \color{RubineRed}{\xmark} & \color{RubineRed}{\xmark} & \color{RubineRed}{\xmark} & \color{RubineRed}{\xmark} \\
    DCQCN & \color{ForestGreen}{\cmark} & \color{ForestGreen}{\cmark} & \color{RubineRed}{\xmark} & \color{ForestGreen}{\cmark} \\
    HPCC / SWIFT & \color{RubineRed}{\xmark} & \color{ForestGreen}{\cmark} & \color{ForestGreen}{\cmark} & \color{ForestGreen}{\cmark} \\
    \textbf{\alg{}} (this paper) & \color{ForestGreen}{\cmark} & \color{ForestGreen}{\cmark} & \color{ForestGreen}{\cmark} & \color{ForestGreen}{\cmark} \\
    \bottomrule
\end{tabular}
\caption{Comparison of various approaches. A {\textit{v}} means the method has successfully controlled and prevented congestion in this task, whereas {\textit{x}} presents a failure. As can be seen, \alg{} is the only learning-based method successful in tackling the task of congestion control, and the only method across all compared methods that succeeds in all tasks.}\label{table: comparison}
\end{center}
\end{table*}

To validate our claims, we develop an RL environment, based on a realistic networking simulator, and perform extensive experiments. The simulator, based on OMNeT++ \citep{varga2002omnet++}, emulates the behavior of state of the art hardware deployed in current datacenters: ConnectX-6Dx Network Interface Card (NIC). In addition, we further test the generalization and robustness of the agent by porting it to \textit{real hardware} and evaluating it there. Our experiments show that our method, \algfull{} (\alg), learns a robust policy, in the sense that it is competitive in all the evaluated scenarios, both in simulation and when tested in the real world. Often, it outperforms the current state-of-the-art methods deployed in real datacenters. 

\section{Related Work} \label{sec: related work}

\textbf{Multi-agent RL:} Recent work in games has shown the premise of multi-agent RL, where trained agents have competed with the top human players. These methods have been applied to tasks ranging from board games such as Go \citep{silver2017mastering} and up to high dimensional 3-dimensional multi-player games such as DOTA \citep{berner2019dota} and StarCraft \citep{vinyals2019grandmaster}. Despite these impressive achievements, we find that due to the combination of challenges in our task (partial observability, multi-objectiveness and multi-agent) the methods in these papers can't be applied in tasks such as datacenter congestion control.

\textbf{Hand-tuned congestion control:} Here, we focus on \textit{datacenter} congestion control. Previous work tackled this problem from various angles. \citet{alizadeh2010data} used a TCP-like protocol, increasing the rate until congestion is sensed, and then dramatically decreasing it. \citet{mittal2015timely,kumar2020swift} focused on round-trip latency measurements to react quickly to changes. \citet{zhu2015congestion} used a statistical signal provided by the switch (ECN), and \citet{li2019hpcc} added telemetry information, requiring specialized hardware, yet proves to benefit greatly in terms of reaction times.

\textbf{Optimization-based congestion control:} Although most previous work has focused on hand-tuned algorithmic behavior, two notable mentions have taken an optimization-based approach. \citet{dong2018pcc} presented the PCC-Vivace algorithm, which combines information from fixed time intervals such as bandwidth, latency inflation, and more. As it tackles the problem via online convex optimization, it is stateless; as such, it does not optimize for long-term behavior but rather focuses on the immediate reward (bandit setting). 

PCC-Vivace was then extended in the Aurora system \citep{jay2019deep}. Aurora provides the monitor interval, defined in PCC-Vivace, as a state for a PPO \citep{schulman2017proximal} algorithm. While in this work we focus on a realistic multi-agent simulator, in \citet{jay2019deep} the focus was on a naive single-agent noisy connection. Thus, we observe that applying the PPO algorithm to our setting, as was done in \citet{jay2019deep}, results in stability and convergence issues (see \cref{table: comparison}). Hence, comparing to \citet{jay2019deep}, our novelty is twofold. (1) We consider a realistic setting that is used within real datacenters to evaluate congestion control algorithms prior to deployment, and (2) rising from the issues highlighted in this task, we present a novel on-policy deterministic policy gradient method that is capable of quickly converging to a satisfying solution.


\section{Networking Preliminaries} 
\label{sec:networking}
In datacenters, traffic contains multiple concurrent data streams transmitting at high rates. The servers, also known as \emph{hosts}, are interconnected through a topology of \emph{switches}. A directional connection between two hosts that continuously transmits data is called a \emph{flow}. We assume, for simplicity, that the path of each flow is fixed.

Each host can hold multiple flows whose transmission rates are determined by a \emph{scheduler}. The scheduler iterates in a cyclic manner between the flows, also known as round-robin scheduling. Once scheduled, the flow transmits a burst of data. The burst's size generally depends on the requested transmission rate, the time it was last scheduled, and the maximal burst size limitation.

A flow's transmission is characterized by two primary values. \emph{Bandwidth}: the average amount of data transmitted, measured in Gbit per second; and \emph{latency}: the time it takes for a packet to reach its destination. \emph{Round-trip-time (RTT)} measures the latency of source$\rightarrow$destination$\rightarrow$source. While 
the latency is often the metric of interest, most systems are only capable of measuring RTT. 

\section{Congestion Control}\label{sec: cc}

\emph{Congestion} occurs when multiple flows cross paths, transmitting data through a single congestion point (switch or receiving server) at a rate faster than the congestion point can process. In this work, we assume that all connections have equal transmission rates, as typically occurs in most datacenters. Thus, a single flow can saturate an entire path by transmitting at the maximal rate. 

Each congestion point in the network has an inbound buffer enabling it to cope with short periods where the inbound rate is higher than it can process. As this buffer begins to fill, the time (latency) it takes for each packet to reach its destination increases. When the buffer is full, any additional arriving packets are dropped.

\subsection{Objective}\label{cc: objective}

CC can be seen as a multi-agent problem. Assuming there are N flows, this results in N CC algorithms (agents) operating simultaneously. Assuming all agents have an infinite amount of traffic to transmit, their goal is to optimize the following metrics (where $\uparrow$/$\downarrow$ mean higher/lower is better, respectively):

\begin{enumerate}
    \item Switch bandwidth utilization ($\uparrow$) -- the \% from maximal transmission rate.
    
    \item Packet latency ($\downarrow$) -- the amount of time it takes for a packet to travel from the source to its destination.
    
    \item Packet-loss ($\downarrow$) -- the amount of data (\% of maximum transmission rate) dropped due to congestion.
    
    \item Fairness ($\uparrow$) -- a measure of similarity in the transmission rate between flows sharing a congested path. We consider $\frac{\min_\text{flows} BW}{\max_\text{flows} BW} \in [0, 1]$.
\end{enumerate}

These objectives may be contradictory. Minimizing latency often comes at the expense of maximizing throughput. Hence, multi-objective schemes present a Pareto-front \citep{liu2014multiobjective} for which optimality w.r.t. one objective may result in sub-optimality of another. However, while the metrics of interest are clear, the agent does not necessarily have access to signals representing them. For instance, fairness is a metric that involves all flows, yet the agent is unaware of how many other transmissions are active and what data they transmit. The agent only observes signals relevant to the flow it controls. As such, it is impossible for a flow to obtain an estimate of the current fairness in the system. Instead, we reach fairness by setting each flow's individual target adaptively, based on known relations between its current RTT and rate. More details on this are given in Sec.~\ref{sec: method}.

\section{Reinforcement Learning Preliminaries}\label{sec: rl}

We model the task of congestion control as a multi-agent partially-observable Markov decision process (POMDP) with multiple objectives and continuous actions, where all agents share the same policy. Each agent observes statistics relevant to itself and does not observe the entire global state.

A POMDP is defined as the tuple $(\mathcal{O}, \mathcal{S}, \mathcal{A},P,R)$ \citep{puterman1994markov,spaan2012partially}. An agent interacting with the environment at state $\state \in \mathcal{S}$ observes an observation $o(\state) \in \mathcal{O}$. After observing $o$, the agent selects a continuous action $\action \in \mathcal{A}$. In a POMDP, the observed state does not necessarily contain sufficient statistics for determining the optimal action. After performing an action, the environment transitions to a new state $\state'$ based on the transition kernel $P(\state' | \state, \action)$ and receives a reward $r(\state, \action) \in R$.

Let $\Pi$ be the set of stationary deterministic policies on $\mathcal{A}$, i.e., if $\pi\in\Pi$ then $\pi: \mathcal{O}\rightarrow \mathcal{A}$.  In this work, we focus on the \textit{average reward} performance metric, also known as the gain of the policy $\pi$, $\rho^\pi(\state) \equiv \lim_{T \rightarrow \infty} \frac{1}{T} \E^\pi[\sum_{t=0}^T r(\state_t,\action_t)\mid \state_0=\state]$, where $\E^\pi$ denotes the expectation w.r.t. the distribution induced by $\pi$. The goal is to find a policy $\pi^*,$ yielding the optimal gain $\rho^*$, i.e., for all $\state\in \mathcal{S}$, $\pi^*(o(\state)) \in \argmax_{\pi\in \Pi} \rho^\pi (\state)$ and the optimal gain is $\rho^{*}(\state) = \rho^{\pi^*}(\state)$.

\subsection{Reinforcement Learning For Congestion Control}
\label{sec: method}

The agent, a congestion control algorithm, controls the data transmission rate at the source. Specifically, the algorithm runs within the network-interface-card (NIC). At each decision point, the agent observes statistics correlated with the specific flow it controls. The agent then acts by determining a new transmission rate for that flow and observes the outcome of this action.
We define the four elements in $(\mathcal{O}, \mathcal{A},P,R)$ (\cref{sec: rl}).

\textbf{Observations.} The agent can only observe information relevant to the flow it controls. In this work, we consider the flow's transmission rate and the RTT measurement.

\textbf{Actions.} The optimal transmission rate depends on the number of agents simultaneously interacting in the network and on the network itself (bandwidth limitations and topology). As such, the optimal transmission rate will vary greatly across scenarios. To ensure the agent is agnostic to the specifics of the network and can easily generalize, we define the next transmission rate $\text{rate}_{t+1}$ as a multiplication of the previous rate with the action. I.e., $\text{rate}_{t+1} = \action_t \cdot \text{rate}_t$, where in our experiments $\action_t \in [0.8, 1.2]$.

\textbf{Transitions.} The transition $\state_t \rightarrow \state_{t}'$ depends on the dynamics of the environment and on the frequency at which the agent is polled to provide an action. Here, the agent acts (is asked to provide an updated transmission rate) once an RTT packet is received. This is similar to the definition of a monitor interval by \citet{dong2018pcc}, but while they considered fixed time intervals, we consider event-triggered (RTT) intervals.

\textbf{Reward.} As the task is a multi-agent partially observable problem, the reward must be designed such that there exists a single fixed-point equilibrium. 

Based on \citet{appenzeller2004sizing}, a good approximation of the RTT inflation ($\text{RTT-inflation} = \frac{\text{RTT}}{\text{base-RTT}}$) in a bursty system, where all flows transmit at the ideal rate, behaves like $\sqrt{N}$, where $N$ is the number of flows. In this case, the combined transmission rate of all flows saturates the congestion point, the system is on the verge of congestion, and the major latency increase is due to the packets waiting in the congestion point's buffer. This latency is orders of magnitude higher than the empty-system routing latency. As such, we can assume that all flows sharing a congested path will observe a similar RTT inflation. We define 
\begin{equation}
    r_t^i = -\left( \text{\bf{target}} - \frac{\text{RTT}^i_t}{\text{base-RTT}^i} \cdot \sqrt{\text{rate}^i_t} \right)^2 \,,
\end{equation}    
where {\bf target} is a constant value shared by all flows, $\text{base-RTT}^i$ is defined as the RTT of flow $i$ in an empty system, and $\text{RTT}^i_t$ and $\text{rate}^i_t$ are respectively the RTT and transmission rate of flow $i$ at time $t$. $\frac{\text{RTT}^i_t}{\text{base-RTT}^i}$ is also called the RTT inflation of agent $i$ at time $t$. The ideal reward is obtained when $\textbf{target} = \frac{\text{RTT}^i_t}{\text{base-RTT}^i} \cdot \sqrt{\text{rate}^i_t}$. Hence, when the \textbf{target} is larger, the ideal operation point is obtained when $\frac{\text{RTT}^i_t}{\text{base-RTT}^i} \cdot \sqrt{\text{rate}^i_t}$ is larger. As increasing the transmission rate increases network utilization and thus the observed RTT, the two grow together. Such an operation point is less latency sensitive (RTT grows) but enjoys better utilization (higher rate).  As \cref{prop: reward is good} shows, maximizing this reward results in a fair solution.

\begin{proposition}\label{prop: reward is good}
    The fixed-point rate (solution) for all $N$ flows sharing a congested path is $\frac{\text{max rate}}{N}$.
\end{proposition}

Informally, the optimal reward for all agents is 0. An agent for which $\frac{\text{RTT}^i_t}{\text{base-RTT}^i} \cdot \sqrt{\text{rate}^i_t} > target$ needs to reduce the transmission rate, which in turn will also reduce the RTT. On the other hand, an agent below the target will act in the opposite direction. As all agents sharing the same congestion point observe approximately the same RTT, the fixed point solution is a fair solution. A formal proof is provided in the supplementary material, in addition to experiments showing how the target affects the behavior.

\section{Our Approach: Analytic  Deterministic Policy Gradient}\label{sec: impl}
In this section, we present the intricate combination of challenges arising in our setup. We explain why existing popular approaches are expected to fail, as we indeed observe and show later in experiments. We then introduce our algorithm that leverages the unique properties of the problem to overcome those challenges. 

\textbf{The challenge.}
We address three challenges rising when learning in multi-agent partially-observable domains. The first is \textit{non-stationarity}. When multiple agents are trained in parallel while interacting in a shared   environment, it becomes non-stationary from the point of view of each agent, as other agents continually change. As a result, one is restricted to on-policy methods to ensure agents are trained on relevant data. Even had the environments changed slowly due to slow learning rate, off-policy methods were still not applicable. Such methods require access to the policies of other agents, which are out of reach in our case.

Second, \textit{partial-observability} hinders value-function estimation. Specifically, policy gradient methods that utilize value functions require direct access to states rather than observations to generate correct gradient estimations \citep{azizzadenesheli2018policy}[Theorem~3.2]. Instead, we shall directly use episodic reward trajectories as done in REINFORCE \citep{williams1992simple}. 

The third challenge is \textit{instability} of stochastic policies. While REINFORCE avoids value-function estimation, it requires the policy to be stochastic. However, stochastic policies in our multi-agent partially-observable setup lead to highly unstable behavior. The multiple agents operate at the same time with the common goal of fairness. Thus, it is essential that they stabilize together to an equilibrium. We observe empirically that this is achievable only with deterministic policies. This also explains the failure of REINFORCE in our experiments later. Such instability  was also observed for PPO by \cite{touati2020stable}.  

The combination of the above three challenges creates a unique set of limitations for a learning algorithm. Namely, it should be on-policy, should not depend on value-function estimation, and needs to support deterministic policies. We now propose an efficient approach that combines all these properties. It is achievable thanks to access to the derivative of the reward function.

\textbf{Our algorithm.} 
To generate deterministic policies, as a first choice one might consider Deterministic Policy Gradient \citep[DPG]{silver2014deterministic}. However, it relies on value function estimation, as do its successors such as DDPG \citep{lillicrap2015continuous}. 
Instead, we work around this by directly estimating the gradient via derivation of the reward function. 

For $s\in {\mathcal S}$, the \algfull{} is defined as 
\begin{align*}
    &\nabla_{\theta} \rho^{\pi_\theta} (s) = \nabla_{\theta}  \lim_{T \rightarrow \infty} \frac{1}{T} \E \left[ \sum_{t=0}^T r\big(o(\state_t), \pi_\theta(o(\state_t))\big) \right] \\
    &= \lim_{T \rightarrow \infty} \frac{1}{T}\E \left[ \sum_{t=0}^T \nabla_{\action} r(o(\state_t), \action)|_{\action=\action_t} \cdot \nabla_{\theta} \pi_\theta(o(\state_t)) \right] \, . \numberthis \label{eqn: on policy gradient}
\end{align*}
Similarly to DPG, it is on-policy. Moreover, despite the partial observability, the gradient estimation is unbiased since it relies on rollouts \citep{azizzadenesheli2018policy}[Eq.~(2)].

The gradient estimator used here is different from common estimators  \citep{sutton2000policy,silver2014deterministic} as it requires access to $\nabla_{\action} r(o(\state_t), \action)$.

\begin{claim}\label{claim: approximate gradient}
    The following is an analytical approximation of the deterministic gradient
    \begin{align*}
        \nabla_{\theta} \rho^{\pi_\theta} (\state) \approx \Bigg[  &\lim_{T \rightarrow \infty} \frac{1}{T} \sum_{t=0}^{T} \Big(\text{{\bf target}} \numberthis \label{eqn: approximate gradient}  \\
        &-  \text{rtt-inflation}_t  \cdot  \sqrt{\text{rate}_t} \Big) \Bigg] \nabla_\theta \pi_\theta (o(\state)) \, .
    \end{align*}
\end{claim}
in the supplementary material, we provide an extensive derivation of \cref{claim: approximate gradient}.

\begin{table*}[t]
    \centering
    \begin{tabular}{l|ccc|ccc|ccc|ccc}
        \multirow{2}{*}{\textbf{}} & \multicolumn{3}{c}{\textbf{128 to 1}} & \multicolumn{3}{c}{\textbf{1024 to 1}} & \multicolumn{3}{c}{\textbf{4096 to 1}} & \multicolumn{3}{c}{\textbf{8192 to 1}} \\
         & SU & FR & QL & SU & FR & QL & SU & FR & QL & SU & FR & QL \\
        \midrule
        Aurora  & 
        \multicolumn{3}{c}{\failed} & \multicolumn{3}{c}{\failed} & \multicolumn{3}{c}{\failed} & \multicolumn{3}{c}{\failed} \\
        PPO & 1 & 26 & 3 & \multicolumn{3}{c}{\failed} & \multicolumn{3}{c}{\failed} & \multicolumn{3}{c}{\failed} \\
        REINFORCE & 51 & 100 & 3 & \textbf{74} & \textbf{70} & \textbf{7} & 53 & 45 & 22 & \multicolumn{3}{c}{\failed} \\ 
        DCQCN & \textbf{100} & \textbf{56} & \textbf{11} & \textbf{100} & \textbf{50} & \textbf{13} & \textbf{95} & \textbf{65} & \textbf{12} & \textbf{95} & \textbf{64} & \textbf{12} \\ 
        HPCC & \textbf{83} & \textbf{96} & \textbf{5} & 59 & 48 & 27 & \multicolumn{3}{c}{\failed} & \multicolumn{3}{c}{\failed} \\ 
        SWIFT & 97 & 94 & 26 & \textbf{89} & \textbf{96} & \textbf{27} & \textbf{88} & \textbf{85} & \textbf{77} & \multicolumn{3}{c}{\failed} 
        \\
        \alg{} (ours) & \textbf{92} & \textbf{95} & \textbf{8} & \textbf{90} & \textbf{70} & \textbf{15} & 91 & 44 & 26 & 92 & 29 & 42 \\
        \bottomrule  
    \end{tabular}
    \caption{\textbf{Many-to-one} test results. Numerical comparison of \alg{} (our method) with various baselines. We color the tests that failed (extensive periods of packet loss) in red. As the task has multiple objectives, Bold face denotes results that are not Pareto-dominated (\cref{def: domination}).}
    \label{tab: many-to-one comparison}
\end{table*}

\section{Experiments} \label{sec: experiments}
We evaluate our approach in two domains. First, in a  simulated domain, using a hardware emulator  built on top of OMNeT++ \citep{varga2002omnet++}. Then, the trained agent is also deployed on real hardware and tested in a real environment.

\textbf{Emulated settings:} We focus on 3 major benchmarks:\\ 
\textbf{(1) Many-to-one: $\mathbf{N \rightarrow 1}$}. $N$ senders transmit data through a single switch to a single receiver. We evaluate the agents on $2^i \rightarrow 1$, for $i \in \{7, 10, 12, 13\}$. The exact configuration is presented in the supplementary material.\\
\textbf{(2) All-to-all:} Multiple servers transmitting data to all other servers. Given $N$ servers, there are $N$ congestion points (switches). Data sent towards server $i$ routes through switch port $i$. This synchronized traffic causes high system load.\\
\textbf{(3) Long-short:}
This scenario evaluates how each agent reacts to changes. A single flow (the `long' flow) transmits an infinite amount of data, while several short flows randomly interrupt it with a short data transmission. The goal is to test how fast the long flow reacts and reduces its transmission rate and how fast the short flows increase their transmission rate. Once the short flows finish transmitting, the long flow should quickly recover to the full line rate. We follow the process from interruption until full recovery. An example of ideal behaviors is presented in the supplementary material.

\textbf{Evaluation setup:} To show our method \emph{generalizes to unseen scenarios}, motivating the use in the real world, we split the scenarios to train and test sets. We \textbf{train} the agents only in the many-to-one domain, on the following scenarios simultaneously: $2 \rightarrow 1, 4 \rightarrow 1,$ and $8 \rightarrow 1$. \textbf{Evaluation} (test) is performed on many-to-one, all-to-all, and long-short scenarios. We provide an extensive overview of the training process, including the technical challenges of the asynchronous CC task, in the supplementary material.

\textbf{Compared Baselines.}
We compare with three RL benchmarks \textbf{(1)~REINFORCE} \citep{williams1992simple} \textbf{(2)~PPO} \citep{schulman2017proximal}, \textbf{(3)~Aurora} \citep{jay2019deep} and three rule-based approaches \textbf{(4)~DCQCN} \citep{zhu2015congestion}, \textbf{(5) HPCC} \citep{li2019hpcc} and \textbf{(6) SWIFT} \citep{kumar2020swift}. For RL-based methods, we used ``official" implementations and for rule-based methods we used in-house implementations that are currently deployed in datacenters. Additional details are provided in the supplementary material.

\textbf{Evaluation metrics:} We report three metrics: \textbf{(1) SU:} Switch Utilization, measured in \%, higher is better, \textbf{(2) FR:} Fairness, defined as $\frac{min rate \cdot 100}{max rate}$, higher is better, \textbf{(3) QL:} Queue Latency, measured in $\mu$ seconds, lower is better. 

In our experiments, bold face denotes results that are not dominated (in the Pareto sense). 
\begin{definition}[Domination]\label{def: domination}
    Result A dominates B if for all metrics $m \in M$ $A(m) \approx B(m)$ and there exists at least one metric $m$ such that $A(m) \gg B(m)$.
    Based on feedback from datacenter managers, we determined that $||A(m) - B(m)|| \leq 5$ to be similar $\forall m$.
\end{definition}

\subsection{Experiments With Simulated Data}\label{sec: sim experiments}
Simulating a networking environment is hard to do correctly. The behavior may be sensitive to hardware limitations, and should take into account packet loss that may occur due to other agents interacting with the network.

To test our approach, we used a hardware emulator built on top of OMNeT++ \citep{varga2002omnet++}. OMNet++ is a sophisticated simulator that is used by the largest data centers to benchmark congestion control algorithms prior to deployment. It accurately emulates hardware devices, including limitations of parallel compute by on-board CPU. While no simulation is perfect, this simulator minimizes the difference between the simulated environment and the real world.

\begin{table}[t]
    \centering
    \label{tab: all-to-all comparison}
    \begin{tabular}{l|ccc|ccc}
        \multirow{2}{*}{\textbf{}} & \multicolumn{3}{c}{\textbf{4 hosts}} & \multicolumn{3}{c}{\textbf{8 hosts}} \\
         & SU & FR & QL & SU & FR & QL \\
        \midrule
        DCQCN & \textbf{89} & \textbf{93} & \textbf{4} & 87 & 66 & 5 \\
        HPCC & 71 & 18 & 3 & \textbf{69} & \textbf{60} & \textbf{3} \\
        SWIFT & \textbf{93} & \textbf{100} & \textbf{10} & \textbf{92} & \textbf{100} & \textbf{11} \\
        {\alg{}} & \textbf{94} & \textbf{77} & \textbf{6} & \textbf{94} & \textbf{97} & \textbf{8} \\
        \bottomrule
    \end{tabular}
    \caption{\textbf{All-to-all} test results. In these tests, none of the algorithms exhibited packet loss.}
\end{table} 

\textbf{Many-to-one:} \cref{tab: many-to-one comparison} gives the results for the many-to-one scenario. Competing methods are competitive when the number of flows and interactions are small. However, with more flows, all methods aside from \alg{} and DCQCN have extensive periods of packet loss and high latency. Across all tasks, previous RL methods (REINFORCE, PPO and Aurora) fail, either converging to extremely safe behavior or constantly transmitting at full-line-rate entirely ignoring latency and packet loss. This validates our arguments that as these methods rely on strong assumptions, they are unfit for many real-world tasks.

\begin{figure*}[t]
    \centering
    \textbf{2 to 1}\\
    \begin{subfigure}[b]{0.192\linewidth}
        \centering
        \includegraphics[width=\linewidth]{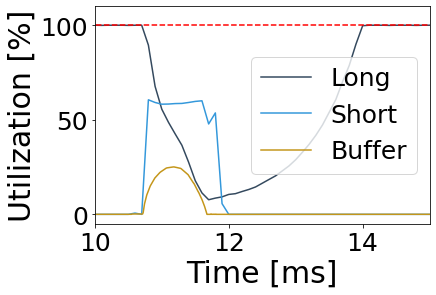}
        \caption{\alg}
        \label{subfig : rl long short 2}
    \end{subfigure}
    \begin{subfigure}[b]{0.161\linewidth}
        \centering
        \includegraphics[width=\linewidth]{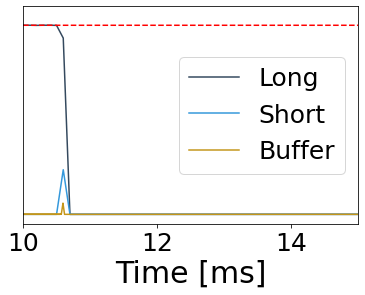}
        \caption{DCQCN}
        \label{subfig : dc2qcn long short 2}
    \end{subfigure}
    \begin{subfigure}[b]{0.161\linewidth}
        \centering
        \includegraphics[width=\linewidth]{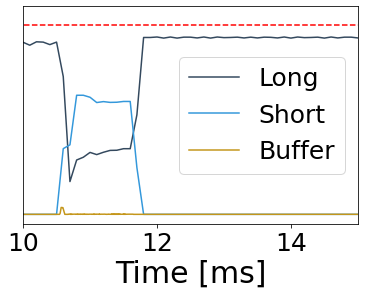}
        \caption{HPCC}
        \label{subfig : hpcc long short 2}
    \end{subfigure}
    \begin{subfigure}[b]{0.161\linewidth}
        \centering
        \includegraphics[width=\linewidth]{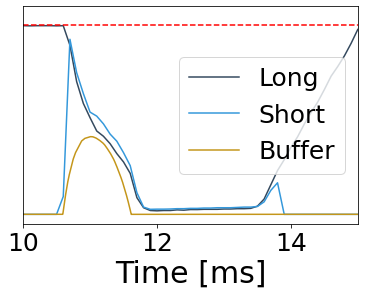}
        \caption{SWIFT}
        \label{subfig : swift long short 2}
    \end{subfigure}\\
    
    \textbf{8 to 1}\\
    \begin{subfigure}[b]{0.196\linewidth}
        \centering
        \includegraphics[width=\linewidth]{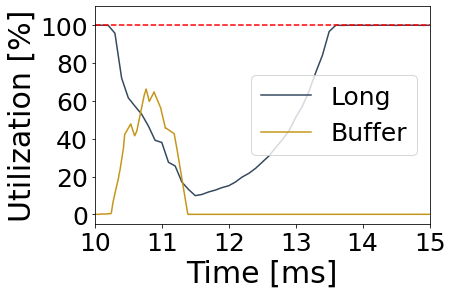}
        \caption{\alg}
        \label{subfig : rl long short 8}
    \end{subfigure}
    \begin{subfigure}[b]{0.161\linewidth}
        \centering
        \includegraphics[width=\linewidth]{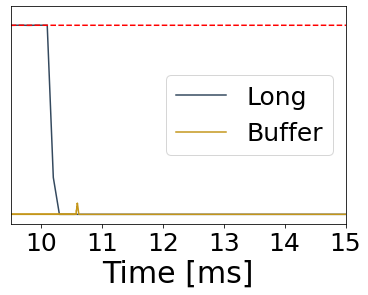}
        \caption{DCQCN}
        \label{subfig : dc2qcn long short 8}
    \end{subfigure}
    \begin{subfigure}[b]{0.161\linewidth}
        \centering
        \includegraphics[width=\linewidth]{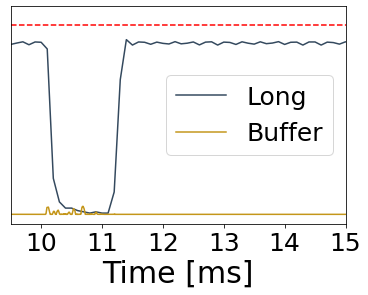}
        \caption{HPCC}
        \label{subfig : hpcc long short 8}
    \end{subfigure}
    \begin{subfigure}[b]{0.161\linewidth}
        \centering
        \includegraphics[width=\linewidth]{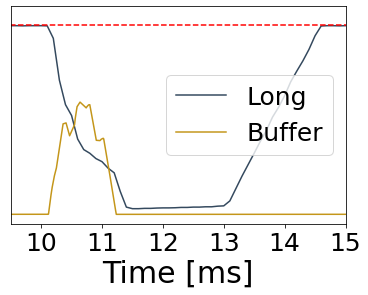}
        \caption{SWIFT}
        \label{subfig : swift long short 8}
    \end{subfigure}
    \caption{ \textbf{Long Short} test results. The goal is to recover fast, but also avoid packet loss. Higher buffer utilization means higher latency. A fully utilized buffer (100\% utilization of the 5MB allocated) leads to packet loss. In these tests, none of the algorithms encountered packet loss. The top row (\cref{subfig : rl long short 2,subfig : dc2qcn long short 2,subfig : hpcc long short 2,subfig : swift long short 2}) presents the results of a long-short test with 2 flows, and the bottom row (\cref{subfig : rl long short 8,subfig : dc2qcn long short 8,subfig : hpcc long short 8,subfig : swift long short 8}) presents a test with 8 flows. We plot the bandwidth utilization of the long flow and the buffer utilization in the switch. Recovery time is measured as the time it takes the long flow to return to maximal utilization. As can be seen, in both scenarios, DCQCN does not recover within a reasonable time. In addition, in HPCC, the long flow does not reach 100\% utilization, even when there are no additional flows.}
    \label{fig: long short}
\end{figure*}

\textbf{All-to-all:} Each method provides a different trade-off balancing between utilization, fairness and latency -- see \cref{tab: all-to-all comparison}. Similarly to the many-to-one scenario, ADPG is capable of generalizing and exhibits competitive behavior.

\textbf{Long-short:} \cref{fig: long short} depicts performance for the long-short scenario, numerical results are in the supplementary material.
Here, algorithms are quantified based on how quickly they react to changes. \alg{} was not trained with this scenario, but performs well. 
Here, HPCC was fastest to react. This is expected since it was specifically designed to handle  long-short scenarios \citep{li2019hpcc}. Unlike  HPCC, \alg{} achieves 100\% utilization before and after the interruption and recovers faster than both SWIFT and DCQCN (which fails in this scenario).

\textbf{Summary:}  We find  that \alg{} learns a robust policy. Although in certain tasks the baselines marginally outperformed it, \alg{} always obtained competitive performance. In addition, \alg{} is the only learning-based method capable of successfully solving the task and thus controlling the network congestion. As \alg{} was trained only in low-scale many-to-one scenarios, it highlights the ability of our method to generalize and learn a robust behavior (successful behavior within higher scale and diverse test scenarios), that we believe will perform well in a real datacenter.

\subsection{Experiments In The Real-World}\label{sec: real world}

Going beyond the simulated environment, we deployed trained agents in real hardware devices.
As existing hardware lacks the accelerators required for fast inference, the reaction time (forward pass) was on the order of $~400 \mu sec$. This was slower than used in our simulated experiments, which were designed to simulate future hardware with $\bigO(1 \mu sec)$ reaction time. Importantly, the baselines we compare with do not rely on DNN inference, thus react within $\bigO(1 \mu sec)$.

\begin{table}[t]
    \centering
    \label{tab: real many-to-one comparison}
    \begin{tabular}{l|c|c|c}
         & \multicolumn{1}{c}{{2 to 1}} & \multicolumn{1}{c}{{1024 to 1}} & \multicolumn{1}{c}{{2000 to 1}} \\
        \midrule
        DCQCN & \textbf{89}  $\pm$ 0.1 & \textbf{86}  $\pm$ 1.3 & \textbf{78}  $\pm$ 0.6 \\
        HPCC & \textbf{84} $\pm$ 0.2 & 68 $\pm$ 0.1 & 72 $\pm$ 0.1 \\
        SWIFT & \textbf{89} $\pm$ 0.0 & \textbf{82} $\pm$ 0.2 & \textbf{83} $\pm$ 0.5 \\
        \alg{} & \textbf{86}  $\pm$ 1.2 & \textbf{82}  $\pm$ 3.9 & 72  $\pm$ 3.2 \\
        \bottomrule
    \end{tabular}
    \caption{\textbf{Real-world, Many-to-one} test results, comparing switch utilization \%.}
\end{table}

We deployed our trained agent from \cref{sec: sim experiments} after performing quantization and porting code to native C. Results are in \cref{tab: real many-to-one comparison}. Even though the ADPG agent was trained under the assumption of a much faster reaction time, it performs on-par with other methods, 
outperforms HPCC in the many-to-one test and is competitive to DCQCN in the low-scale setting. We expect dramatic improvement with faster hardware (future generation).

\section{Summary}

AI centric algorithms have the ability to leverage direct interaction with the datacenter in order to optimize performance. We presented the fundamental challenges this task presents when using reinforcement learning tools. These challenges prevent popular algorithms from successfully solving the task. We proposed an on-policy deterministic policy gradient algorithm and presented a way to use domain knowledge to obtain an analytical gradient estimation.

Experiments in a simulated environment, a realistic OMNeT++ network simulator commonly used to benchmark CC algorithms for deployment in real datacenters, show that our method \alg{} successfully addresses this task. More generally, they demonstrate the efficacy and generalization capabilities of an RL approach which is in contrast to the hand-crafted algorithms that currently dominate the field of CC. While some baselines achieve outstanding performance in specific evaluation scenarios, they catastrophically fail in others. In contrast, \alg{} learned a robust policy that performed well across all scenarios we tested, and often obtained the best results. We also show that \alg{} generalizes to unseen domains and is capable of operating at a large-scale without incurring packet loss, a setting where most competing methods fail.

Finally, we deployed our trained agent in real hardware. Current hardware suffers from high latency, an issue that we expect to solve with the next generation NICs. Regardless, our method was capable of successfully obtaining competitive behavior, thus increasing our confidence in its applicability when combined with future low-latency hardware.

While there are yet many challenges on the path of applying RL in the real world, we believe these results to be a promising start. While the baseline algorithms have gone through years of meticulous hand tuned optimization, \alg{} was capable of quickly obtaining competitive behavior both in simulation and in the real world.

\bibliography{bibliography}

\onecolumn
\appendix

\section{Simulator}\label{apndx: sim}

The simulator attempts to model the network behavior as realistically as possible. The task of CC is a multi-agent problem, there are multiple flows running on each host (server) and each flow is unaware of the others. As such, each flow is a single agent, and 4096 flows imply 4096 concurrent agents.

Each agent is called by the CC algorithm to provide an action. The action, whether continuous or discrete, is mapped to a requested transmission rate. When the flow is rescheduled, it will attempt to transmit at the selected rate. Calling the agent (the triggering event) occurs each time an RTT packet arrives.

Agents are triggered by spontaneous events rather than at fixed time intervals; this makes the simulator asynchronous. Technically speaking, as the simulator exposes a single step function, certain agents might be called upon more times than others.

While the action sent to the simulator is for the current state $s_t$, in contradiction to the standard GYM environments, state $s_{t+1}$ is not necessarily from the same flow as $s_t$. Due to the asyncronous nature of the problem, the simulator provides the state which corresponds to the next agent that receives an RTT packet.

To overcome this, we propose a `KeySeparatedTemporalReplay', a replay memory that enables storing asynchronous rollouts separated by a key (flow). We utilize this memory for training our method and Aurora (PPO), who both require gradient calculation over entire rollouts.

\subsection{Many to one tests}

In the many to one tests, each configuration combines a different number of hosts and flows. For completeness we provide the exact mapping below:

\begin{table}[H]
    \centering
    \caption{Many to one experiment mapping}
    \label{tab: many to one mapping}
    \begin{tabular}{c|c|c}
    Total flows & Hosts & Flows per per host \\
    2 & 2 & 1 \\
    4 & 4 & 1 \\
    16 & 16 & 1 \\
    32 & 32 & 1 \\
    64 & 64 & 1 \\
    128 & 64 & 2 \\
    256 & 32 & 8 \\
    512 & 64 & 8 \\
    1024 & 32 & 32 \\
    2048 & 64 & 32 \\
    4096 & 64 & 64 \\
    8192 & 64 & 128 
    \end{tabular}
\end{table}

\subsection{Computational Details}

The agents were trained on a standard i7 CPU with 6 cores and a single GTX 2080. The training time (for 200k steps) took 2-3 hours. The major bottleneck was the evaluation times. We evaluated the agents in a many-to-one setting with a very high number of flows (up to 8k). The more flows, the longer the test. In the python version, on this system it took approximately 2 days to evaluate the agent throughout 2 simulated seconds. However, on an optimized C implementation, the same evaluation took 20 minutes.

\section{Fixed-Point Proof}
\label{sec: fixed point proof}

\begin{proposition}
    The fixed-point solution for all $N$ flows sharing a congested path is a transmission rate of $\frac{1}{N}$.
\end{proposition}

The proof relies on the assumption that all flows sharing a congested path observe the same rtt inflation. Although the network topology affects the packet path and thus the latency, this latency is minimal when compared to the queue latency of a congested switch.

\begin{proof}
    ~\\
    The maximal reward is obtained when all agents minimize the distance $||\text{rtt-inflation} \cdot \sqrt{rate} - \textbf{target}||$. There are two stationary solutions (1) $\text{rtt-inflation} \cdot \sqrt{rate} < \textbf{target}$ or (2) $\text{rtt-inflation} \cdot \sqrt{rate} = \textbf{target}$.
    
    As flows can always reduce the transmission rate (up to 0) and $\text{rtt-inflation} \propto \sqrt{rate}$. A solution where $\text{rtt-inflation} \cdot \sqrt{rate} > \textbf{target}$ is not stable.
    
    
    We analyze both scenarios below and show that a stable solution at that point is also fair.
    
    \begin{enumerate}
        \item $\text{rtt-inflation} \cdot \sqrt{rate} < \textbf{target}$. The value is below the target. Minimizing the distance to the target means maximizing the transmission rate. A stable solution below the target is obtained when the flows are transmitting at full-line rate (can't increase the rate over 100\%) and yet the rtt-inflation is low (small or no congestion). As all flows are transmiting at 100\% this solution is fair.
        
        \item $\text{rtt-inflation} \cdot \sqrt{rate} = \textbf{target}$. For any $i, j$ sharing a congested path, we assume that $\text{rtt-inflation}_i = \text{rtt-inflation}_j$, this is a reasonable assumption in congested systems as the RTT is mainly affected by the latency in the congestion point. As all flows observe $\text{rtt-inflation} \cdot \sqrt{rate} = \textbf{target}$, we conclude that if $\text{rtt-inflation} \cdot \sqrt{rate} = \textbf{target}$ then $\sqrt{rate}_i = \sqrt{rate}_j = \frac{1}{N}, \forall i,j$.
    \end{enumerate}
\end{proof}

\section{Training Curves}\label{apndx: experiments}

In this section of the appendix, we expand on additional methods that did not prevail as well as our algorithm.

\subsection{Aurora}
We begin with Aurora \citep{jay2019deep}. Aurora is similar to PCC-RL in how it extracts statistics from the network. A monitor-interval (MI) is defined as a period of time over which the network collects statistics. These statistics, such as average drop rate, RTT inflation, and more, are combined into the state provided to the agent.

However, Aurora focused on the task of single-agent congestion control. As they considered internet congestion control (as opposed to datacenter congestion control), their main challenge was handling jitters (random noise in the network resulting in packet loss even at network under-utilization).

An additional difference is that Aurora defines a naive reward signal, inspired by PCC-Vivace \citep{dong2018pcc}:
\begin{equation*}
    r = a \cdot BW - b \cdot RTT - c \cdot DROP RATE \enspace , a,b,c \geq 0
\end{equation*}

\begin{figure}[H]
    \centering
    \includegraphics[width=0.5\textwidth]{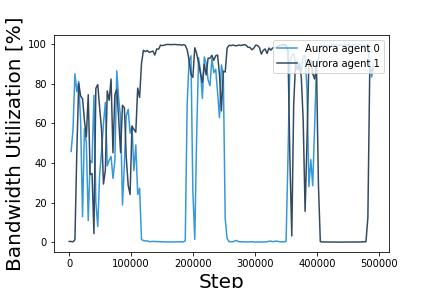}
    \caption{\textbf{Aurora training:} two hosts with a single flow per host. The flows are unable to converge to a stable equilibrium.}
    \label{fig:aurora training}
\end{figure}

We observe in \cref{fig:aurora training} that Aurora is incapable of converging in the simulated environment. We believe this is due to the many challenges the real world exhibits, specifically partial observability and the non-stationarity of the opposing agents.

\subsection{PPO}

PCC-RL introduces a deterministic on-policy policy-gradient scheme that utilizes specific properties of the reward function.

It is not immediately clear why such a scheme is important. As such, we compare to PPO trained on our raw reward function
\begin{equation*}
    r = - (\textbf{target} -  \text{rtt-inflation} \cdot  \sqrt{\text{rate}})^2
\end{equation*}

We present two versions of PPO. (1) A continuous action space represented as a stochastic Gaussian policy $a \sim \mathcal{N} (\mu, \sigma) \, , \, \mu \in [0.8, 1.2]$ (as is common in continuous control tasks such as MuJoCo \citep{todorov2012mujoco,schulman2017proximal}). (2) A discrete action space represented as a stochastic discrete policy (softmax) where $a \in \{0.8, 0.95, 1, 1.05, 1.1, 1.2 \}$.

\begin{figure}[H]
    \centering
    \begin{subfigure}[b]{0.49\linewidth}
        \centering
        \includegraphics[width=\linewidth]{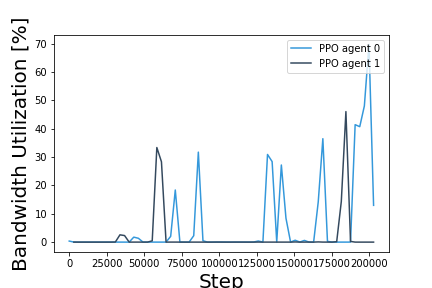}
        \caption{Continuous}\label{ppo cont}
    \end{subfigure}
    \begin{subfigure}[b]{0.49\linewidth}
        \centering
        \includegraphics[width=\linewidth]{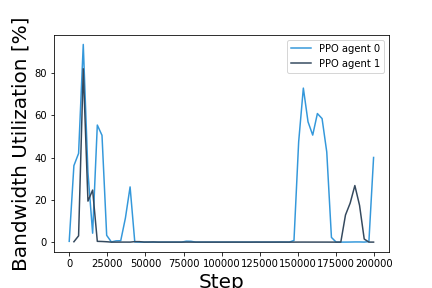}
        \caption{Discrete}\label{ppo disc}
    \end{subfigure}
    \caption{\textbf{PPO training:} both the continuous (\cref{ppo cont}) and discrete (\cref{ppo disc}) versions of the PPO algorithm are unable to learn, even with our target-based reward signal.}
    \label{fig:ppo}
\end{figure}

\subsection{PCC-RL}

\begin{figure}[H]
    \centering
    \includegraphics[width=0.5\textwidth]{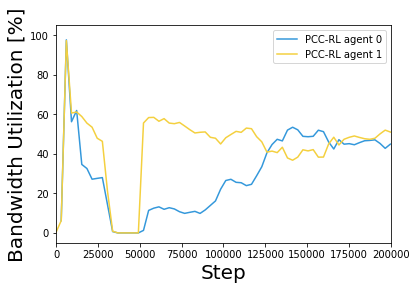}
    \caption{\textbf{PCC-RL training:} the agents quickly converge to a region of the fair equilibrium.}
    \label{fig:pcc rl training}
\end{figure}

Finally, we present the training curves of PCC-RL. As can be seen, \cref{fig:pcc rl training}, PCC-RL quickly converges to a region of the fixed-point stable equilibrium.

In addition, below we present the performance of a trained PCC-RL agent during a 4 to 1 evaluation test.

\begin{figure}[H]
    \centering
    \includegraphics[width=0.5\linewidth]{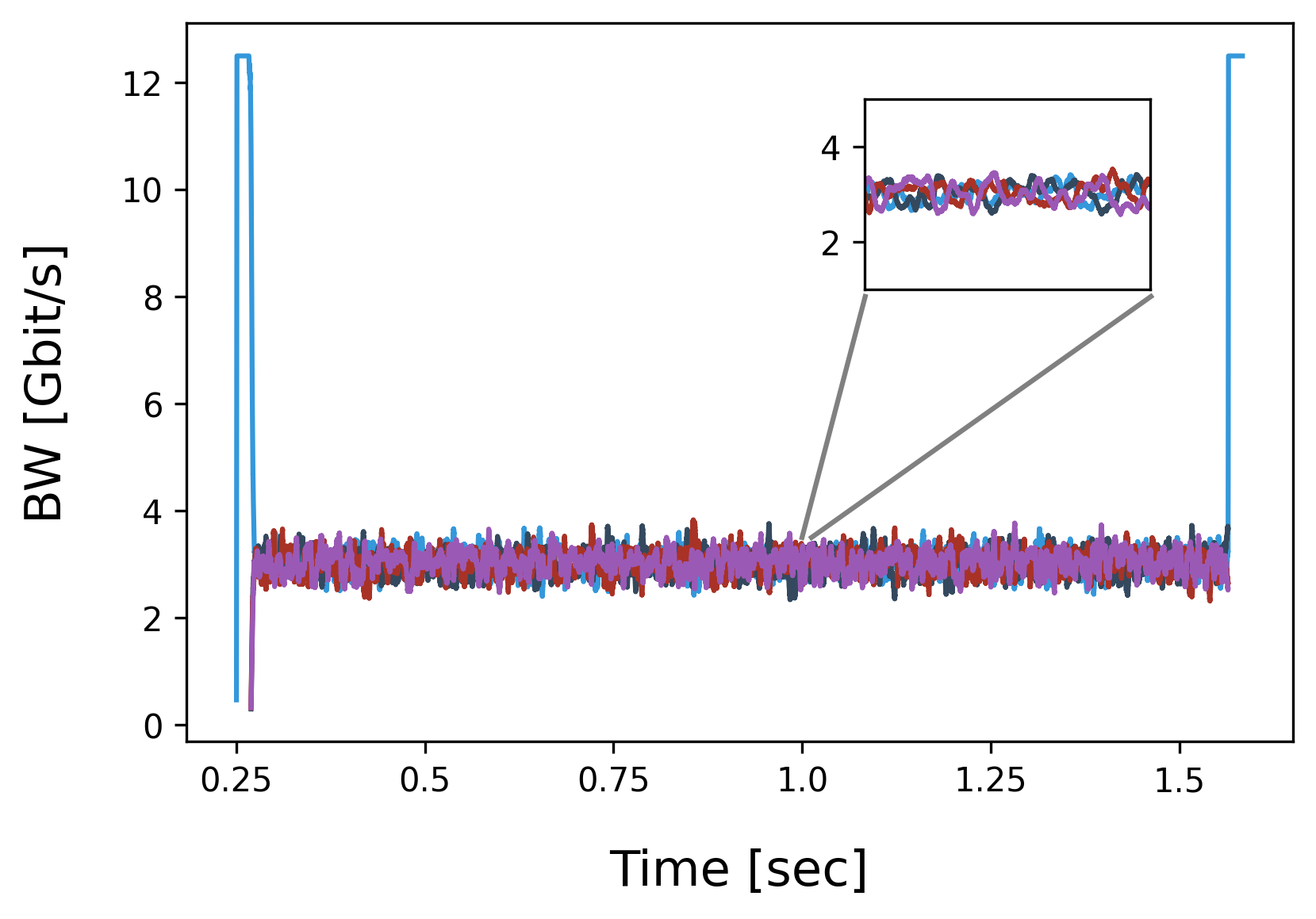}
    \caption{Four PCC-RL agents controlling 4 flows, each running on a different host, and marked with a different line color. Only a single flow is active at the beginning and end of the simulation. The congestion point is limited to $100 [GByte/s] = 12.5 [Gbit/s]$. As seen, the flows quickly converge to a region of the fair transmission rate ($\frac{12.5}{4}$).}
    \label{fig: 4 to 1 packet flow}
\end{figure}

\section{Quantization}\label{apndx: quantization}

A major challenge in CC is the requirement for real-time decisions. Although the network itself is relatively small and efficient, when all computations are performed in int8 data type, the run-time can be dramatically optimized.

To this end, we test the performance of a trained PCC-RL agent after quantization and optimization in C. Following the methods in \citet{wu2020integer}, we quantize the network to int8. Combining int8 operations requires a short transition to int32 (to avoid overflow), followed by a de-quantization and re-quantization step.

We present the results in \cref{tab: quantization comparison}. The quantized agent performs similarly to an agent trained on a \emph{loose} target (improved switch utilization at the expense of a slightly higher latency). These exciting results show that a quantized agent is capable of obtaining similar performance to the original agent. This is a major step forward towards deployment in live datacenters.

\begin{table}[H]
    \centering
    \caption{\textbf{Quantization many to one:} Comparison of the original Python PCC-RL agent with an optimized and quantized C version.}
    \label{tab: quantization comparison}
    \begin{tabular}{c||c|c||c|c||c|c}
        \multirow{2}{*}{\textbf{Flows}}& \multicolumn{2}{c}{\textbf{Switch Utilization}} & \multicolumn{2}{c}{\textbf{Fairness}} & \multicolumn{2}{c}{\textbf{Queue Latency}} \\
         & \textbf{Original} & \textbf{Quantized} & \textbf{Original} & \textbf{Quantized} & \textbf{Original} & \textbf{Quantized} \\
        \hline
        \textbf{2} & 96.9 & 99.91 & 0.99 & 1.00 & 5.2 & 8.29 \\
        \hline
        \textbf{4} & 96.9 & 99.86 & 0.98 & 1.00 & 5.4 & 8.59 \\
        \hline
        \textbf{16} & 95.1 & 99.61 & 0.99 & 0.99 & 6.2 & 9.46 \\
        \hline
        \textbf{32} & 95.6 & 99.22 & 0.86 & 0.99 & 7.0 & 9.90 \\
        \hline
        \textbf{64} & 93.3 & 99.06 & 0.96 & 0.98 & 7.0 & 9.72 \\
        \hline
        \textbf{128} & 92.5 & 98.57 & 0.94 & 0.96 & 8.0 & 10.94 \\
        \hline
        \textbf{256} & 91.4 & 98.02 & 0.91 & 0.90 & 9.3 & 12.85 \\
        \hline
        \textbf{512} & 90.4 & 97.54 & 0.86 & 0.86 & 11.3 & 16.26 \\
        \hline
        \textbf{1024} & 90.2 & 97.10 & 0.74 & 0.75 & 14.7 & 20.92 \\
        \hline
        \textbf{2048} & 90.5 & 96.74 & 0.56 & 0.61 & 20.3 & 27.67 \\
        \hline
        \textbf{4096} & 91.3 & 96.65 & 0.46 & 0.43 & 27.7 & 36.87 \\
        \hline
        \textbf{8192} & 92.8 & 96.79 & 0.28 & 0.29 & 40.0 & 48.40
    \end{tabular}
\end{table}

\section{Long Short Details}\label{apndx: longshort}

\begin{table}[H]
    \centering
    \caption{\textbf{Long-short:} The results represent the time it takes from the moment the \emph{short} flows interrupt and start transmitting until they finish and the \emph{long} flow recovers to full line rate transmission. DC2QCN does not recover fast enough and has thus failed the high scale recovery tests. We present the recovery time \textbf{RT} ($\mu$sec), drop rate \textbf{DR} (Gbit/s) and the bandwidth utilization of the long flow \textbf{LBW} (\%).}
    \label{tab: long-short comparison}
    \begin{tabular}{c||c|c|c||c|c|c||c|c|c||c|c|c}
        \multirow{2}{*}{\text{Algorithm}} & \multicolumn{3}{c}{\textbf{2 flows}} & \multicolumn{3}{c}{\textbf{128 flows}} & \multicolumn{3}{c}{\textbf{1024 flows}} & \multicolumn{3}{c}{\textbf{2048 flows}} \\
         & RT & DR & LBW & RT & DR & LBW & RT & DR & LBW & RT & DR & LBW \\
        \hline\hline
        \textbf{PCC-RL} & \textbf{6e-7} & \textbf{0} & \textbf{97} & \textbf{8e-4} & \textbf{0} & \textbf{94} & 3e-2 & 1.1 & 62 & 3e-2 & 2.7 & 56 \\
        \hline
        \textbf{DC2QCN} & 3e-2 & 0 & 63 & 5e-2 & 0 & 40 & \multicolumn{3}{c}{\cellcolor{red!40}-} & \multicolumn{3}{c}{\cellcolor{red!40}-} \\
        \hline
        \textbf{HPCC} & 3e-5 & 0 & 90 & 1e-2 & 0.4 & 75 & 2e-2 & 1.1 & 72 & 3e-2 & 1.8 & 62 \\
        \hline
        \textbf{SWIFT} & 1e-3 & 0 & 97 & 1e-2 & 0.3 & 85 & \textbf{1e-2} & \textbf{1.2} & \textbf{83} & \textbf{2e-2} & \textbf{2.1} & \textbf{72}
    \end{tabular}
\end{table}

\end{document}